\pdfoutput=1
\def\year{2017}\relax
\documentclass[letterpaper]{article} 
\usepackage{aaai17}  
\usepackage{times}  
\usepackage{helvet}  
\usepackage{courier}  
\usepackage{url}  
\usepackage{graphicx}  
\usepackage{amsmath,amssymb}
\usepackage{amsthm}
\usepackage{adjustbox}
\usepackage{array}
\usepackage{xr}

\newcolumntype{R}[2]{%
>{\adjustbox{angle=#1,lap=\width-(#2)}\bgroup}%
l%
<{\egroup}%
}
\newcommand*\rot{\multicolumn{1}{R{75}{1em}}}
\newcommand{\pddl}[1]{{\small \textsf{#1}}}

\theoremstyle{definition}
\newtheorem{theorem}{Theorem}

\newtheorem{definition}{Definition}
\newcommand{\maintain}[1]{{x^{maintain}_{f,#1}}}
\newcommand{\preadd}[1]{{x^{preadd}_{f,#1}}}
\newcommand{\predel}[1]{{x^{predel}_{f,#1}}}
\newcommand{\add}[1]{{x^{add}_{f,#1}}}
\newcommand{\del}[1]{{x^{del}_{f,#1}}}
\newcommand{\sat}[1]{{x^{sat}_{#1,t}}}

\newcommand{\op}{{y_{o,t}}}

\newcommand{\Def}{{{\rm Def}_P(u)}}
\newcommand{\Ext}{{{\rm Ext}_P(u)}}
\newcommand{\Int}{{{\rm Int}_P(u)}}
\newcommand{\IS}{{{\rm IS}(u,r)}}
\newcommand{\SCC}{{{\rm SCC}(u)}}

\begin{document}
%
\title{Axioms in Model-based Planners}
\author{
		Shuwa Miura {\rm and} Alex Fukunaga\\
		Graduate School of Arts and Sciences\\ 
		The University of Tokyo\\
		miura-shuwa@g.ecc.u-tokyo.ac.jp, fukunaga@idea.c.u-tokyo.ac.jp\\
}
\maketitle
\begin{abstract}
Axioms can be used to model derived predicates in domain-
independent planning models. Formulating models which use
axioms can sometimes result in problems with much smaller
search spaces and shorter plans than the original model.
Previous work on axiom-aware planners focused solely on state-space search planners.
We propose axiom-aware planners based on answer set programming and integer proramming. We evaluate them on PDDL domains with axioms and show that they can exploit additional expressivity of axioms.
\end{abstract}

\section{Introduction}
Currently, in the most commonly studied classical planning models, all changes to the world are the direct effects of some operator.
However, it is possible to model some effects as indirect effects which can be inferred from a set of basic state variables. Such \emph{derived predicates} can be expressed in modeling languages such as PDDL and formalisms such as SAS+ as \emph{axioms}, which encode logical rules defining how the derived predicates follow from basic variables.
Planners have supported  various forms of derived predicates since relatively early systems \cite{MannaW87,Barrett95}, 
and PDDL 
has supported axioms which specify derived predicates as a logic program with negation-as-failure semantics since version 2.2 \cite{EdelkampH2004}

Consider, for example, the well-known single-agent puzzle game \pddl{Sokoban},
in which the player pushes stones around in a maze. 
The goal is to push all the stones to their destinations.
The standard PDDL formulation of \pddl{Sokoban} used in the International Planning Competition(IPC) 
consists of two kinds of operators, \pddl{push} and \pddl{move}.
\pddl{push} lets the player push a box in one direction, while \pddl{move} moves the player into an unoccupied location.

\citeauthor{ivankovic2015code} (\citeyear{ivankovic2015code}) proposed a new formulation of Sokoban with axioms and showed that this leads to a problem with a smaller search space and shorter plan \cite{ivankovic2015optimal}.
They remove the \pddl{move} operators entirely, and introduce  axioms to check whether the player can reach a box to push it.
The reformulated \pddl{push} operators now have a derived predicate 
\pddl{reachable(loc)} instead of \pddl{at-player=loc} as their precondition.
The values of the derived predicates are determined by the following axioms:
		\begin{enumerate}
			\item \pddl{reachable(loc)} $\leftarrow$ \pddl{at-player=loc}
			\item \pddl{reachable(loc)} $\leftarrow$  \pddl{reachable(from)}, \pddl{clear(loc)}, \pddl{connected(from,loc)}
		\end{enumerate}
Intuitively, the first axiom means that the current location of the player is reachable.
The second axiom means a location next to a reachable location is also reachable.
With axioms,
 the search space only has the transitions caused by \pddl{push} operators, resulting in smaller search space and shorter plan.

Previous work on derived predicates and axioms for planning has focused on the advantages with respect expressivity (compactness) of domain modeling using axioms, \cite{thiebaux2005defense}, 
as well as forward state-space search algorithms which are aware of axioms \cite{ColesS07,GereviniSS11,ivankovic2015optimal}. 

One class of approaches to planning translates planning problem instances to instances of other domain-independent modeling frameworks such as SAT \cite{kautz1992planning} and Integer Programming (IP) \cite{vossen1999use}.
We refer to planners based on such translation-based approaches \emph{model-based planners}.
While previous work focused solely on axiom-aware state-space-based planners, to our knowledge, little or no work has been done on PDDL domains with axioms to evaluate model-based planners
 A standard approach to model-based  planning is to translate a planning problem instance into a $k$-step SAT/IP/CSP model,  
where a feasible solution to the $k$-step model corresponds to a solution to the original planning instance with $n < k$ ``steps''.
To our knowledge, no previous work has evaluated \emph{axiom-aware, model-based approaches} to planning.

 We propose two axiom-aware model-based planners called ASPlan and IPlan, which are based on answer set programming (ASP) and integer programming (IP) respectively.
 Answer set programming (ASP) is a form of declarative programming based on answer set semantics of logic programming, and thus 
 is a natural candidate for integrating axioms.
 Early attempts to apply ASP to planning \cite{subrahmanian1995relating} and \cite{DimopoulosNK97} predated PDDL, and were not evaluated on large sets of benchmarks.
To our knowledge, the first ASP-based planner compatible with  PDDL is  {\it plasp} \cite{gebser2012gearing}. 
However, {\it plasp} does not handle axioms.
 We developed ASPlan based on {\it plasp}, but used the PDDL to SAS+ translator from FastDownward \cite{helmert2006fast} to make it compatible with larger sets of IPC domains including domains with axioms.
IPlan is an IP-based planner based on Optiplan \cite{briel2005optiplan}.
 We integrated axioms into IPlan by using the ASP to IP translation method by \cite{liu2012answer}.

To our knowledge, ASPlan and IPlan are the first model-based planners which handle axioms.
We evaluate ASPlan and IPlan on an extensive set of benchmark domains with axioms, and show that additional expressivity of axioms benefit model-based planners as well.
The rest of this paper is structured as follows. 
We first review background material including normal logic problems, SAS+ with axioms, and  the search semantics for model-based planning (Section \ref{sec:preliminaries}).
Then, we describe ASPlan, our axiom-aware answer set programming based planner (Section \ref{sec:asplan}) and IPlan, our axiom-aware, IP-based planner (Section \ref{sec:iplan}).
Section \ref{sec:experiments} presents our experimental evaluation of ASPlan and IPlan on an extensive set of domains with axioms, including IPC domains as well as domains used in previous work on planning with axiom-aware forward search planning \cite{ivankovic2015optimal,Ghosh2015,kominis2015beliefs}.
We conclude with a summary of our contributions (Section \ref{sec:conclusion}).

\section{Preliminaries}
\label{sec:preliminaries}
\subsection{Normal Logic Problem}
\label{sec:nlp}
 We introduce normal logic problems, adopting the notations used in \cite{liu2012answer}.
\begin{definition}
	A {\it normal logic problem} (NLP) $P$ consists of rules of the form
	\begin{equation}
		\label{eq:rule}
		a \leftarrow b_1, ... , b_m, {\rm not} c_1, ... , {\rm not} c_n.
	\end{equation}
	where each $a$, $b_i$, $c_j$ is a ground atom.
\end{definition}
 Given a rule $r \in P$, we denote the head of rule $a$ by $H(r)$, the body $\{ b_1,..., b_m, {\rm not} c_1,..., {\rm not} c_n \}$ as $B(r)$,
  the positive body literals $\{ b_1,..., b_m \}$ as $B^+(r)$ and the negative body literals $\{ {\rm not} c_1,..., {\rm not} c_n \}$ as $B^-(r)$.
We use $At(P)$ for the set of atoms which appear in $P$.

A set of atoms $M$
satisfies an atom $a$  
if $a \in M$ and a negative literal ${\rm not} a$ if
 $a \not \in M$, denoted $M \vDash a$ and $M \vDash {\rm not} a$, respectively;
$M$ satisfies a set of literals $L$, denoted $M \vDash L$, if it satisfies
 each literal in $L$; $M$ satisfies a rule $r$, denoted
 $M \vDash r$, if $M \vDash H(r)$ whenever $M \vDash B(r)$. 
 A set of atoms $M$ is a model of $P$, denoted $M \vDash P$, if $M$
 satisfies each rule of $P$.

 An answer set of a program is defined through the concept of reduct.
\begin{definition}
 For a normal logic program $P$ and a set of atoms  $M$, the {\it reduct} $P^M$ is defined as 
	\begin{equation}
		P^M = \{H(r) \leftarrow B^+(r) \mid r \in P, B^-(r) \cap M = \emptyset \}.
	\end{equation}
\end{definition}

 \begin{definition}[\citeauthor{gelfond1988stable} \citeyear{gelfond1988stable}]
	A model $M$ of a normal logic program $P$ is an {\it answer set} iff
	$M$ is the minimal model of $P^M$.
\end{definition}

Consider a program $P_1$ with the following rules:
\begin{align}
	(r_1) \quad a &\leftarrow {\rm not } b. \\
	(r_2) \quad b &\leftarrow {\rm not} a. \\
	(r_3) \quad c &\leftarrow a.
\end{align} 
$M_1=\{a,c\}$ and $M_2=\{b\}$ are the answer sets of the program $P_1$ since they are the minimal model of $P_1^{M_1}$ and $P_1^{M_2}$ respectively.

We are particularly interested in {\it locally stratified programs}, 
which disallow negation through recursions \cite{Przymusinski1988OnTD}.
\begin{definition}
	\label{def:stratified}
	 A normal logic program $P$ is {\it locally stratified} if and only if there is a mapping $l$ from $At(P)$ to 
	 $\{1,...,\lvert At(p) \rvert\}$ such that:

\begin{itemize}
	\item for every rule $r$ with $H(r) = a$ and every $b \in B^+(r)$ , 				$l(b) \leq l(a)$
	\item for every rule $r$ with $H(r) = a$ and every $c \in C^-(r)$, 				$l(c) < l(a)$
\end{itemize}
\end{definition}

Consider a program $P_2$ with the following rules:
\begin{align}
	(r_4) \quad a &\leftarrow {\rm not } b. \\
	(r_5) \quad b &\leftarrow c. \\
	(r_6) \quad c &\leftarrow b.
\end{align} 
	$P_2$ is locally stratified since there exists a mapping $l$ with $l(a)=2$ and $l(b)=l(c)=1$ satisfying Definition \ref{def:stratified}. Note that there is no mapping $l$ for $P_1$ due to the negative recursions through the rules $r_1$ and $r_2$.

With stratifications the unique model, called a {\it perfect model} can be computed by a stratified fixpoint procedure \cite{apt1988towards}. It is known that the perfect model for a locally stratified program coincides with the unique answer set of the program \cite{eiter2009answer}. Indeed, $P_2$ has the unique model $M=\{a\}$.

\subsection{SAS+ and Axioms}
We adopt the definition of SAS+ or finite domain representation (FDR) used in \citeauthor{helmert2009concise} (\citeyear{helmert2009concise}) and \citeauthor{ivankovic2015optimal} (\citeyear{ivankovic2015optimal}).
\setlength{\thickmuskip}{0mu} 
\begin{definition}
	An SAS+ problem $\Pi$ is a tuple $(\allowbreak V \allowbreak ,\allowbreak U \allowbreak ,\allowbreak A \allowbreak ,\allowbreak O,\allowbreak I,\allowbreak G \allowbreak )$ where

\noindent -   $V$ is a set of \emph{primary variables}. Each variable $v_i$ has a finite domain of values $D(v_i)$.

\noindent - $U$ is a set of \emph{secondary variables}. Secondary variables are binary and do 
not appear in operator effects.
Their values are determined by axioms after each operator execution.

		\noindent - $A$ is a set of rules of the form (\ref{eq:rule}).
Axioms and primary variables form a locally stratified logic program.
At each state, axioms are evaluated to derive the values of secondary variables, resulting in an extended state.
We denote the result of evaluating a set of axioms $A$ in a state $s$ as $A(s)$ and simply call it a state when there is no confusion.
Note that $A(s)$ is guaranteed to be unique due to the uniqueness of the model for locally stratified logic programs.
\label{def:sas+axioms}

\noindent - $O$ is a set of operators. Each operator $o$ has a precondition (pre($o$)), which consists of variable assignments of the form $v_i=x$ where $x \in D(v_i)$. We abbreviate $v_i=1$ as $v_i$ when we know the variable is binary. 
		An operator $o$ is applicable in a state $s$ iff 
		$s$ satisfies pre($o$).

		Each operator $o$ also has a set of effects (eff($o$)).
		Each effect $e \in $eff($o$) consists of a tuple (cond($e$), affected($e$)) where 
		cond($e$) is a possibly empty variable assignment and affected($e$) is a single variable-value pair. 
		Applying an operator $o$ with an effect $e$ to a state $s$ where cond($e$) is satisfied results in a state ($o(s)$)
		where affected($e$) is true.
		Although the original definition \cite{helmert2009concise} does not specify this, we assume that conflicting effects, which assign different values to the same 
		variable, never get triggered.

		cost($o$) is the cost associated with the operator $o$.

\noindent - $I$ is an initial assignment over primary variables.

\noindent - $G$ is a partial assignment over variables that specifies the goal conditions.

\end{definition}

A solution (\emph{plan}) to  $\Pi$ is an applicable sequence of operators $o_0,...,o_n$ that maps $I$ into a state where $G$ holds.
\subsection{$\forall$ vs. sequential ($seq$-) semantics}
\label{sec:forall-sequential}
 A standard, \emph{sequential search strategy} for a model-based planner first generates a 1-step model (e.g., SAT/IP model), and attempts to solve it. If it has a solution, then the system terminates. Otherwise, a 2-step model is attempted, and so on \cite{kautz1992planning}. 
Adding axioms changes the semantics of a ``step'' in the $k$-step model.
	As noted by \citeauthor{DimopoulosNK97} (\citeyear{DimopoulosNK97}) and Rintanen et al (\citeyear{RintanenHN06}),  most model-based planners, including the base IPlan/ASPlan planners we evaluate below, use $\forall$ semantics,  where each ``step'' in a $k$-step model consists of a set of operators which are independent of each other and can, therefore, be executed in parallel.
In contrast, \emph{sequential} semantics ($seq$-semantics) adds exactly 1 operator at each step in the iterative, sequential search strategy.
In general,  solving a problem using $\forall$ semantics is faster than with $seq$-semantics, since $\forall$ semantics significantly decreases the number of iterations of the sequential search strategy.
For the domains with axioms, however, we add a constraint which restricts the number of operators executed at each step to 1, imposing $seq$-semantics.
This is because 
a single operator can have far-reaching effects on derived variables, and 
establishing independence with respect to all derived variables affected by multiple operators is non-trivial. 

\section{ASPlan}
\label{sec:asplan}
We describe our answer set programming based planner ASPlan.
ASPlan adapts the encoding of {\it plasp} \cite{gebser2012gearing} to the multi-valued semantics of SAS+.
While {\it plasp} directly encodes PDDL to ASP, ASPlan first
obtains the grounded SAS+ model from PDDL using the FastDownward translator, and then encodes the SAS+ to ASP.
Using the FastDownward translator makes it easier to handle more advanced features like axioms and conditional effects.

\subsection{Baseline Implementation}
We translate a planning instance to an answer set program with $k$ steps.
Having $k$-steps means that we have $k+1$ states or ``layers'' to consider.
	The notation here adheres to the ASP language standard\footnote{\url{https://www.mat.unical.it/aspcomp2013/ASPStandardization}}.
\begin{equation}
	\label{eq:asplan:const}
	\text{ const n = k. step 1..n. layer 0..n.}
\end{equation}
For each $v_i=x \in I$, we introduce the following rule
which specifies the initial state.
\begin{equation}
	\label{eq:asplan:init}
	\text{holds(f($v_i$,$x$),0)}
\end{equation}
Likewise, for each $v_i=x \in G$, we have the following rule.
\begin{equation}
	\label{eq:asplan:g}
	\text{goal(f($v_i$,$x$))}
\end{equation}
The next rule (\ref{eq:asplan:goal}) makes sure that all of the goals are satisfied at the last step.
\begin{equation}
	\label{eq:asplan:goal}
	\text{$\leftarrow$ goal(F), not holds(F,n)}
\end{equation}

For each operator $o \in O$ and each $v_i=x$ and $v_j=y$ in $o$'s preconditions and effects respectively, we introduce the following rules.
\begin{equation}
	\label{eq:asplan:operator}
	\text{ demands($o$,f($v_i$,$x$)). add($o$,f($v_j$,$y$)).}
\end{equation}
Rules (\ref{eq:asplan:demand}) and (\ref{eq:asplan:add}) require that applied operators' preconditions and effects must be realized.
\begin{equation}
	\label{eq:asplan:demand}
	\text{ $\leftarrow$ apply(O,T), demands(O,F), not holds(F,T-1), step(T).}
\end{equation}
\begin{equation}
	\label{eq:asplan:add}
	\text{ holds(F,T) $\leftarrow$ apply(O,T), add(O,F), step(T)}
\end{equation}

Inertial axioms (unchanged variables retain their values) are represented by rules (\ref{eq:asplan:changed})-(\ref{eq:asplan:inertial}).
An assignment $v=x$ to a SAS+ variable is mapped to an atom f($v$,$x$).
\begin{equation}
	\label{eq:asplan:changed}
	\text{ changed(X,T) $\leftarrow$ apply(O,T), add(O,f(X,Y)), step(T)}
\end{equation}
\begin{multline}
	\label{eq:asplan:inertial}
	\text{ holds(f(X,Y),T) $\leftarrow$ holds(f(X,Y),T-1), step(T),} \\
		\text{inertial(f(X,Y)), not changed(X,T)}
\end{multline}
Note that every primary variable $v$ is marked as inertial with the following rule.
\begin{equation}
	\label{eq:asplan:mark_inertial}
	\text{inertial(f($v$,$val$))}
\end{equation}

With sequential semantics,
rule (\ref{eq:asplan:sequential}) ensures that only one operator is applicable in each step.
\begin{equation}
	\label{eq:asplan:sequential}
	\text{ 1 \{apply(O,T) : operators(O)\} 1, step(T)}
\end{equation}

On the other hand,
with $\forall$-semantics, (\ref{eq:asplan:all})-(\ref{eq:asplan:all3}) requires that no conflicting operators are applicable at the same step.
\begin{equation}
	\label{eq:asplan:all}
	\text{ 1 \{apply(O,T) : operators(O)\}, step(T)}
\end{equation}
\begin{multline}
	\label{eq:asplan:all2}
		\text{ $\leftarrow$ apply(O,T), apply(O',T), add(O,f(X,Y)),} \\ \text{ demands(O',f(X,Z)), step(T), O != O', Y != Z}
\end{multline}
\begin{multline}
	\label{eq:asplan:all3}
		\text{  $\leftarrow$ apply(O,T), apply(O',T), add(O,f(X,Y)),} \\ \text{ add(O',f(X,Z)), step(T), O != O', Y != Z.}
\end{multline}

Since variables in SAS+ cannot take different values at the same time, we introduce the mutex constraint (\ref{eq:asplan:mutex}).
A mutex is a set of fluents of which at most one is true in every state.

\begin{equation}
	\label{eq:asplan:mutex}
	\text{ $\leftarrow$ holds(f(X,Y),T), holds(f(X,Z),T), Y != Z, layer(T).}
\end{equation}

\subsection{Integrating Axioms}
Integrating axioms to ASPlan is fairly straightforward.
For an axiom $a \leftarrow b_1,...,b_n,not c_1, ..., c_m$,
we introduce the following rule:
\begin{multline}
	\text{ holds(f($a$,1),T) $\leftarrow$ holds(f($b_1$,1),T),...,holds(f($b_n$,1),T),} \\
	\text{ not holds(f($c_1$,0),T),..., not holds(f($c_m$,0),T), step(T).}
\end{multline}
We also need the following constraints to realize negation-as-failure semantics while being compatible with the formulations above.
\begin{equation}
	\text{ holds(f($u$,0),T) $\leftarrow$ not holds(f($u$,1),T), layer(T).}
\end{equation}
\begin{equation}
	\text{ $\leftarrow$ not holds(f($u$,0),T), not holds(f($u$,1),T), layer(T).}
\end{equation}

\subsection{Integrating Conditional Effects}
We describe how to integrate conditional effects into ASPlan.
For each operator $o$ and its effect $e\in$eff($o$), we introduce the following rule
\begin{multline}
	\text{ fired(E,T) $\leftarrow$ apply(A,T),}\\ 
	\text{ holds(f($v_1$,$x$),T), ..., holds(f($v_n$,$z$),T), step(T)}
\end{multline}
where $v_1=x$,...,$v_n=z$ are in cond($e$).

With conditional effects, applying operators does not necessarily mean their effects get triggered.
We replace rules (\ref{eq:asplan:add}) and (\ref{eq:asplan:changed}) with the following rules.
\begin{equation}
	\text{ holds(F,T) $\leftarrow$ fired(E,T), add(E,F), effect(E), step(T)}
\end{equation}
\begin{equation}
	\label{eq:asplan:newchanged}
	\text{ changed(X,T) $\leftarrow$ fired(E,T), add(E,f(X,Y)), effect(E), step(T)}
\end{equation}

\section{IPlan}
\label{sec:iplan}
\subsection{Baseline Implementation}
We describe our baseline integer-programming planner IPlan, which is
based on Optiplan \cite{briel2005optiplan}.
Optiplan, in turn,  extends the state-change variable model \cite{vossen1999use}, and the Optiplan model is defined for an propositional (STRIPS) framework.
IPlan adapts the Optiplan model for the multi-valued SAS+ framework, exploiting the  FastDownward translator \cite{helmert2006fast}.

An assignment $v=x$ to a SAS+ variable is mapped to a fluent $f$.
For all fluents $f$ and time step $t$, Optiplan has the following binary \emph{state change variables}.
$pre_f$, $add_f$ and $del_f$ denote a set of operators that might require, add, or delete $f$ respectively.
Intuitively, state change variables represent all possible changes of fluents at time step $t$.
\begin{itemize}
	\item $\maintain{t} = 1 \text{ iff fluent } f \text{ is propagated in step } t$
	\item $\preadd{t} = 1 \text{ iff } o \in pre_f \setminus del_f \text{ is executed in step } t$
	\item $\predel{t} = 1 \text{ iff } o \in pre_f \cap del_f \text{ is executed in step } t$
	\item $\add{t} = 1 \text{ iff } o \in add_f \setminus pre_f \text{ is executed in step } t$
	\item $\del{t} = 1 \text{ iff } o \in del_f \setminus pre_f \text{ is executed in step } t$
\end{itemize}

For all operators $o$ and time step $t$, Optiplan has the binary operator variables.
\begin{equation*}
	\op = 1 \text{iff operator } o \text{ is executed in period } t
\end{equation*}

Constraints (\ref{eq:iplan:initial}) and (\ref{eq:iplan:initial2}) represent the initial states constraints.
\begin{align}
	\label{eq:iplan:initial}
	\add{0} &= 1 \quad \forall f \in I \\
	\add{0}, \maintain{0}, \preadd{0}  &= 0 \quad \forall f \not \in I
	\label{eq:iplan:initial2}
\end{align}

Constraint (\ref{eq:iplan:goal}) ensures that the goals are satisfied at the last step $T$ ($\sat{f}$ is introduced later).
\begin{equation}
	\label{eq:iplan:goal}
	\sat{f}  \geq1 \quad \forall f \in G, t = T
\end{equation}

For all fluents $f$ and time step $t$, Optiplan has the following constraints to make sure
the state change variables have the intended semantics.
\begin{align}
	\label{eq:iplan:add}
	\sum_{o \in add_f \setminus pre_f} \op &\geq \add{t} \\
	\op &\leq \add{t} \quad \forall o \in add_f \setminus pre_f
\end{align}

\begin{align}
	\sum_{o \in del_f \setminus pre_f} \op &\geq \del{t} \\
	\op &\leq \del{t} \quad \forall o \in del_f \setminus pre_f
\end{align}

\begin{align}
	\sum_{o \in pre_f \setminus del_f} \op &\geq \preadd{t} \\
	\op &\leq \preadd{t} \quad \forall o \in pre_f \setminus del_f
\end{align}

\begin{align}
	\label{eq:iplan:predel}
	\sum_{o \in pre_f \cap del_f} \op &= \predel{t}
\end{align}

Constraints (\ref{eq:iplan:para}) and (\ref{eq:iplan:para2}) restrict certain state changes 
from occurring in parallel.
\begin{align}
	\label{eq:iplan:para}
	\add{t} + \maintain{t} + \del{t} + \predel{t} &\leq 1 \\
	\label{eq:iplan:para2}
	\preadd{t} + \maintain{t} + \del{t} + \predel{t} &\leq 1
\end{align}

Constraint (\ref{eq:iplan:backward}) represents the backward chaining requirement, that is, if a fluent $f$
is true at the beginning of step $t$ then there must have been a change that made $f$ true at step $t-1$, 
or it was already true at $t-1$ and maintained.
\begin{multline}
	\label{eq:iplan:backward}
	\preadd{t} + \maintain{t} + \predel{t} \leq \\ \preadd{t-1} + \add{t-1} + \maintain{t-1} \quad \forall f \in F, t \in 1,...,T
\end{multline}

\subsubsection{New Mutex Constraints}

IPlan augments the Optiplan model with a new set of mutex constraints.
In addition to the above variables and constraints which are from Optiplan, 
IPlan introduces a set of auxiliary binary variables 
$\sat{f}$, which correspond to the value of the fluent $f$ at the time step $t$ and are constrained as follows.
\begin{equation}
	\sat{f} \leq \add{t} + \preadd{t} + \maintain{t}
\end{equation}
\begin{equation}
	\sat{f} \geq \add{t}
\end{equation}
\begin{equation}
	\sat{f} \geq \preadd{t}
\end{equation}
\begin{equation}
	\sat{f} \geq \maintain{t}
\end{equation}

Using $\sat{f}$, mutex constraints can be implemented as follows.\footnote{This mutex constraint was proposed by Horie in an unpublished undergraduate thesis in \cite{Horie2015}.}
For every mutex group $g$ (at most one fluent in $g$ can be true at the same time) found by the FastDownward \cite{helmert2006fast} translator,
IPlan adds the following constraint.
\begin{equation}
	\label{eq:iplan:mutex}
	\sum_{f \in g} \sat{f} \leq 1
\end{equation}
\citeauthor{ghanbari2017encoding}  (\citeyear{ghanbari2017encoding}) used similar mutex constraints for thier CP-based planner.

\subsection{Integrating Axioms}
We describe how to integrate axioms into an IP-based model.
For each time step $t$, axioms form
the corresponding normal logic program (NLP) $P_t$.
The models for $P_t$ correspond to the truth values for the derived variables. 
We translate each NLP $P_t$ to an integer program (IP) using the method by \citeauthor{liu2012answer} (\citeyear{liu2012answer}), and add these linear constraints to the IPlan model.

\subsubsection{Level Rankings}
Translation from a NLP to an IP by \citeauthor{liu2012answer} (\citeyear{liu2012answer}) relies
on characterizing answer sets using {\it level rankings}.
Intuitively, a level ranking of a set of atoms gives an order in which the atoms are derived.
\begin{definition}[\citeauthor{niemela2008stable} \citeyear{niemela2008stable}]
Let $M$ be a set of atoms and $P$ a normal program.
A function $lr : M \rightarrow N$ is a {\it level ranking} of $M$ for $P$ iff for each $a \in M$, there is a rule $r \in P_M$ such that $H(r) = a$ and for every $b \in B^+(r)$, $lr(a) - 1 \geq lr(b)$.
\end{definition}

Note that $P_M$ is the set of support rules, which are essentially the rules applicable under $M$.
\begin{definition}[\citeauthor{niemela2008stable} \citeyear{niemela2008stable}]
For a program $P$ and $I \subset At(P)$, 
	$P_I = \{r \in P \mid I \vDash B(r)\}$ is the set of {\it support rules}.
\end{definition}
The level ranking characterization gives the condition under which a supported model is
an answer set.

\begin{definition}[\citeauthor{apt1988towards} \citeyear{apt1988towards}]
\label{def:supported}
A set of atoms $M$ is a supported model of a program
$P$ iff $M \vDash P$ and for every atom $a \in M$ there is a rule
$r \in P$ such that $H(r) = a$ and $M \vDash B(r)$.
\end{definition}

\begin{theorem}[\citeauthor{niemela2008stable} \citeyear{niemela2008stable}]
Let $M$ be a supported model
of a normal program $P$. Then $M$ is an answer set of $P$ iff
there is a level ranking of $M$ for $P$.
\end{theorem}

Consider, for example, program $P_2$ (from Section \ref{sec:nlp}) and its model $M' = \{b,c\}$.
$M'$ is a supported model for $P_2$ satisfying the Definition \ref{def:supported}.
$M'$, however, is not an answer set of $P_2$ since there is no level ranking for $M'$ with the atoms $b$ and $c$ forming a cycle.

\subsubsection{Components and Defining Rules}
\begin{figure}[htb]
\centering
	\includegraphics[width=0.8\linewidth]{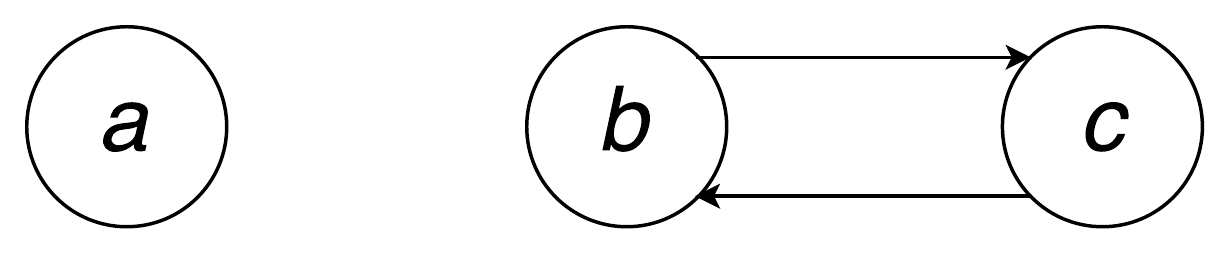}
	\caption{The dependency graph for $P_2$.}
	\label{fig:graph} 
\end{figure}

\citeauthor{liu2012answer} (\citeyear{liu2012answer}) define
a dependency graph of a normal logic program to be used for the translation to IP.
\begin{definition}[\citeauthor{liu2012answer} \citeyear{liu2012answer}]
	The {\it dependency graph} of a program $P$ is a directed graph
	$G$ = $\langle V, E \rangle$ where $V$ = $At(P)$ and $E$ is a set of edges $\langle a, b \rangle$
	for which there is a rule $r \in P$ such that $H(r)$ = $a$ and
	$b \in B^+(r)$.
\end{definition}
For example, the dependency graph for $P_2$ is shown in Figure \ref{fig:graph}.
We use SCC($a$) to denote the strongly connected component (SCC) containing an atom $a$.

\begin{definition}[\citeauthor{liu2012answer} \citeyear{liu2012answer}]
For a program $P$ and an atom $a \in At(P)$,
	respective sets of {\it defining rules}, {\it externally defining rules},
	and {\it internally defining rules} are defined as follows:
\begin{equation}
	{\rm Def}_P(a) = \{r \in P \mid H(r) = a \}
\end{equation}
\begin{equation}
	{\rm Ext}_P(a) = \{r \in {\rm Def}_P(a) \mid B^+(r) \cap {\rm SCC}(a) = \emptyset \}
\end{equation}
\begin{equation}
	{\rm Int}_P(a) = \{r \in {\rm Def}_P(a) \mid B^+(r) \cap {\rm SCC}(a) \neq \emptyset \}
\end{equation}

	The set of {\it internally supporting atoms} is defined as 
\begin{equation}
	{\rm IS}(a, r) = {\rm SCC}(a) \cap B^+(r).
\end{equation}
\end{definition}
For example, for $P_2$, ${\rm Def}_{P_2}(a) = {\rm Ext}_{P_2}(a) = \{ r_4 \}$, ${\rm Def}_{P_2}(b) = {\rm Int}_{P_2}(b) = \{ r_5 \}$ and ${\rm Def}_{P_2}(c) = {\rm Int}_{P_2}(c) = \{ r_6 \}$.

\subsubsection{Translation}
We are now ready to describe how to translate
a normal logic program $P_t$ formed by axioms to linear constraints based on the method by \citeauthor{liu2012answer} (\citeyear{liu2012answer}).
The translation consists of linear constraints developed below.
\begin{enumerate}
	\item For each secondary variable $u \in U$, introduce a binary variable $\sat{u}$.

	\item For each secondary variable $u \in U$, include the following constraint
\begin{equation}
	\sum_{r \in \Def} bd^r_t - |\Def| \cdot \sat{u} \leq 0
\end{equation}
	where $bd^r_t$ is a binary variable for each $r \in \Def$ and step $t$.
	Intuitively, $bd^r_t$ represents whether the body of $r$ is satisfied at step $t$.
	The constraint ensures that when one of the rules defining $u$ ($\Def$) is satisfied, $\sat{u}$ must be true ($\sat{u} = 1$).
		For example, for atom $a$ in $P_2$, this introduces the constraint $bd^{r_4}_t - 1 \cdot \sat{a} \leq 0$.

	\item For each axiom $r \in A$, include the following constraints.
\begin{equation}
	\label{eq:axiom:body}
	\sum_{b \in B^+(r)} \sat{b}  - \sum_{c \in B^-(r)} \sat{c} - |B(r)| \cdot bd^r_t \geq -|B^-(r)|
\end{equation}
\begin{equation}
	\label{eq:axiom:body2}
	\sum_{b \in B^+(r)} \sat{b}  - \sum_{c \in B^-(r)} \sat{c} - bd^r_t \leq |B^+(r)| - 1
\end{equation}
		Constraints (\ref{eq:axiom:body}) and (\ref{eq:axiom:body2}) express the fact that the body of rule $r$ is satisfied iff each literal in $B(r)$ is satisfied.
		For example, for $r_6$ in $P_2$, this introduces the constraints $\sat{b} - 1 \cdot bd^{r_6} \geq - 0$ and $\sat{b} - bd^{r_6}_t \leq 1 - 1$.
	\item For each secondary variable $u \in U$, include the constraint
\begin{equation}
	\sum_{r \in \Ext} bd^r_t  + \sum_{r \in \Int} s^r_t - \sat{u} \geq 0
\end{equation}
	where $s^r_t$ is a binary variable for each $r \in \Int$ and each step $t$.
		Intuitively, the binary variable $s^r_t$ represents whether the respective ranking constraints for the rule $r$ are satisfied in addition to its body. The constraint requires $\sat{u}$ to be true when one of its externally defining rules is satisfied or one of its internally defining rules is satisfied while respecting the ranking constraints.
		For example, for atom $a$ and $c$ in $P_2$, this introduces the constraints $bd^{r_4}_t + 0 - \sat{a} \geq 0$ and $0 + s^{r_6}_t - \sat{c} \geq 0$ respectively.

	\item For each secondary variable $u \in U$ and each $r \in \Int$, include the constraints
\begin{equation}
	bd^r_t - s^r_t \geq 0
\end{equation}
\begin{equation}
	\sum_{b (b \in \IS)} gt^{ub}_t - |\IS| \cdot s^r_t \geq 0
\end{equation}
		where $gt^{ub}_t$ is a binary variable for each
		$b \in \IS$, which represents whether the rank of 
		$u$ is greater than that of $b$.
		For example, for atom $c$ and $r_6$ in $P_2$, this introduces the constraints $gt^{cb}_5 - 1 \cdot s^{r_6}_5 \geq 0$.

	\item For each secondary variable $u \in U$ and each $r \in \Int$, and each $b \in \IS$, include the constraint
\begin{equation}
	\label{eq:axiom:gt}
	z_{a,t} - z_{b,t} - |\SCC| \cdot gt^{ab}_t \geq 1 - |\SCC|
\end{equation}
		where $z_{a,t}$ and  $z_{b,t}$ are 
		integer variables representing level rankings for $a$ and $b$ respectively.
		Constraint (\ref{eq:axiom:gt}) guarantees that if $gt^{ab}_t=1$  then $z_{a,t} \geq z_{b,t}$.
		e.g., for atom $c$ in $P_2$, this introduces the constraint $z_{c,t} - z_{b,t} - 2 
		\cdot gt^{cb}_t \geq 1 - 2$.
\end{enumerate}

In the above constraints, $\sat{u}$ corresponds to 
the values of a secondary variable $u$ at time step $t$.
Since secondary variables only appear in operator preconditions,
we only need to make sure the preconditions of applied operators are satisfied.
\begin{equation}
	y_{u,t} \leq \sat{u} \quad \forall a \in pre_f \setminus del_f \end{equation}

As explained in Section \ref{sec:forall-sequential} , in case of domains with axioms, we need to add the following constraint to restrict the number of operators
executed at each time step to 1.
\begin{equation}
	\label{eq:iplan:sequential}
	\sum y_{a,t} \leq 1
\end{equation}

\section{Experimental Results}
\label{sec:experiments}

All experiments are performed on a Xeon E5-2670 v3, 2.3GHz with 2GB RAM and 5 minute time limit. 
In all experiments, the runtime limits include
all steps of problem solving, including translation/parsing, and search.
We use clingo4.5.4, a state-of-the-art ASP solver \cite{clingo} to solve the ASP models produced by ASPlan.
The IP models produced by IPlan are solved using Gurobi Optimizer 6.5.0, single-threaded.

The rules and constraints used in each of our planner configurations are summarized in Table \ref{tab:rules}.

\subsection{Baseline Evaluation on PDDL Domains without Axioms}

We first evaluated ASPlanS and IPlanS on PDDL domains without axioms to compare them against existing planners.
A thorough comparison of model-based planners is non-trivial because of the multitude of 
combinations possible of translation schemes, solvers, and search strategies.
The purpose of this experiment is to show that the baseline ASPlan and IPlan planners perform reasonably well compared to \emph{similar}, existing model-based planners which (1) use a simple search strategy which iteratively solves $k$-step models, and (2) use models which are solved by ``off-the-shelf'' solver algorithms, i.e., this excludes planners that such as Madagascar \cite{RintanenHN06}, which uses a more sophisticated search strategy  and customized solver algorithm, as well as the flow-based IP approaches \cite{BrielVK08}.

We compared the following: 
(1) {\it plasp}\footnote{{\it plasp} was sourced from \\ (\url{https://sourceforge.net/p/potassco/code/7165/tree/trunk/plasp-2.0/releases}).}
(2) APlanS (ASPlan with seq-semantics) 
(3) IPlanS (IPlan with seq-semantics)
(4) ASPlan (ASPlan with $\forall$-semantics)
(5) IPlan (IPlan with $\forall$-semantics)
(6) SCV (IPlan without the mutex constraints)
(7) TCPP, a state of the art CSP-based planner \cite{ghanbari2017encoding}. 

We used {\it plasp} with the default configurations with sequential semantics. Since {\it plasp} uses incremental grounding, we used iClingo\footnote{iClingo was obtained from \\ (\url{https://sourceforge.net/projects/potassco/files/iclingo/3.0.5/iclingo-3.0.5-x86-linux.tar.gz/download})} \cite{gekakaosscth08a} as an underlying solver. Note that ASPlan could have used incremental grounding as well, but we decided not to for ease of implementation.
As for TCPP, we could not obtain the souce code from the authors as of this writing, so we use the results from the original paper.
We chose the overall-best-performing configuration TCPPxm-conf-sac, which uses \emph{don't care} and mutex constraints proposed in \cite{ghanbari2017encoding}.
The results are shown in Table \ref{tab:no-axioms}.

 ASPlanS dominated {\it plasp} in every domain, indicating that ASPlan is a reasonable, baseline ASP-based solver. This is to be expected, since ASPlanS is based on the {\it plasp} model while utilizing FastDownward tranlator for operator grounding, although ASPlans does not use incremental grounding.
Among $\forall$-semantics solvers ASPlan, IPlan and TCPP, ASPlan and IPlan are highly competitive with TCPP despite the fact that the results for TCPP were obtained with a significantly longer (60min. vs. 5min) runtime limit (on a different machine).
Comparing IPlan and SCV, additional mutex constraints increased coverage in domains such as \pddl{blocks} and \pddl{logistics00}.


\begin{table}[htb]
	\scalebox{0.9}{
		\begin{tabular}{|l|l|l|l|}
		\hline
				 ASPlan
				& ASPlanS
				& IPlan
				& IPlanS \\
			\hline
			(\ref{eq:asplan:const})-(\ref{eq:asplan:mark_inertial})
			,(\ref{eq:asplan:all})-(\ref{eq:asplan:mutex})
			& (\ref{eq:asplan:const})-(\ref{eq:asplan:sequential})
			,(\ref{eq:asplan:mutex})-(\ref{eq:asplan:newchanged})
			& (\ref{eq:iplan:initial})-(\ref{eq:iplan:mutex})
			& (\ref{eq:iplan:initial})-(\ref{eq:iplan:sequential}) \\
			\hline
		\end{tabular}
	}
	\caption{Summary of the rules or constraints used in each planner configuration}
	\label{tab:rules}
\end{table}

\begin{table}[htb]
	\begin{tabular}{l|r|r|r|r||r|r|r|r|}
		\hline
		& \#
		& \rot{plasp}
		& \rot{ASPlanS}
		& \rot{IPlanS}
		& \rot{ASPlan}      
		& \rot{IPlan} 
		& \rot{SCV} 
		& \rot{TCPP}\\

		\hline
		blocks              & 35         & 15              & 18           & 28       & 18     & 28      & 16  & 32  \\
		depot               & 22         & 2               & 2            & 2        & 9      & 11      & 7  & 11  \\
		driverlog           & 20         & 4               & 7            & 4        & 14     & 11      & 11  & 13  \\
		grid                & 5          & 1               & 2            & 1        & 0      & 1       & 0  & 2  \\
		gripper             & 20         & 2               & 2            & 2        & 3      & 4       & 4  & 2  \\
		freecell            & 80         & 6               & 7            & 7        & 1      & 18      & 18 & 4  \\
		logistics-98        & 35         & 2               & 2            & 1        & 12     & 16      & 11  & 9  \\
		logistics00         & 28         & 7               & 7            & 6        & 28     & 28      & 22  & 24  \\
		movie               & 30         & 30              & 30           & 30       & 30     & 30      & 30  & N/A \\
		mprime              & 35         & 24              & 27           & 21       & 11     & 25      & 24  & 26  \\
		mystery             & 30         & 16              & 16           & 13       & 9      & 16      & 13  & 17  \\
		rovers              & 40         & 0               & 4            & 4        & 23     & 21      & 18  & 22  \\
		satellite           & 36         & 3               & 4            & 5        & 11     & 8       & 8  & 7 \\
		zenotravel          & 20         & 6               & 7            & 3        & 13     & 13      & 11  & 12  \\
		\hline
	\end{tabular}
	\caption{Results on domains without axioms with 2GB, 5min limits. For TCPP, the results from \cite{ghanbari2017encoding} 
	on a Intel Xeon 2.60GHz CPU, 4GB, 60 min limit.
	are shown. N/A indicates a lack of the results. \# denotes the number of instances for each domain.
	}
	\label{tab:no-axioms}
	\centering
\end{table}

\subsection{Evaluation on PDDL Domains with Axioms}
We evaluated ASPlan and IPlan on PDDL domains with axioms.

Below, we describe the domains (other than the previously described \pddl{Sokoban}) used in experimental evaluations,\footnote{The benchmarks are available with more detailed descriptions at (\url{https://github.com/dosydon/axiom\_benchmarks}).}

\subsubsection{Verification Domains} 

\citeauthor{Ghosh2015} (\citeyear{Ghosh2015}) proposed a modeling formalism for capturing high level
functional specifications and requirements of reactive control systems.
The formulation consists of two agents, namely the environment which disturbs the system, and the controller, which tries to return the system to a safe state. If there is a sequence of operators for the environment that leads to an unsafe state, the control system has a vulnerability. We used two compiled versions of the formulation: (a) compilation into STRIPS \cite{Ghosh2015},  and (b) compilation into STRIPS with axioms \cite{ivankovic2015optimal}.

\pddl{ACC} is a verification domain for 
Adaptive Cruise Control (ACC), a well known driver assistance feature present in many high
end automotive system which 
is designed to take away the burden of adjusting the speed of the vehicle from the driver, mostly
under light traffic conditions.

The \pddl{GRID} domain is a synthetic planning domain loosely based on cellular automata and
incorporates a parallel depth first search protocol for added variety.
Note that this domain is completely different from the standard IPC \pddl{grid} domain.
To avoid confusion, we denote this verification domain as \pddl{GRID-VERIFICATION}.

\subsubsection{Multi-Agent Domains} 
\citeauthor{kominis2015beliefs} (\citeyear{kominis2015beliefs}) proposed a framework for handling beliefs in multiagent settings, building on the methods for representing beliefs for a single agent. Computing linear multiagent plans for the framework can be mapped to a classical planning problem with axioms. 
We use three of their domains.

\pddl{Muddy Children}, originally from \citeauthor{fagin2004reasoning} (\citeyear{fagin2004reasoning}), is a puzzle in which $k$ out of $n$ children have mud on their foreheads.
Each child can see the other children's foreheads but not their own.
The goal for a child is to know if he or she has mud by sensing the beliefs of the others.
\pddl{Muddy Child} is a reformulation of \pddl{Muddy Children}.

In \pddl{Collaboration through Communication}, the goal for an agent is to know a particular block's location. Two agents volunteer information to each other to accomplish a task faster than that would be possible individually.

In \pddl{Sum}, there are three agents, each with a number on their forehead. It is known that one of the numbers equals the sum of the other two. The goal is for one or two selected agents to know their numbers.
	
\pddl{Wordrooms} is a variant of {\pddl Collaboration through Communication} where two agents must find out a hidden word from a list of n possible words.

\subsubsection{PROMELA}
This is a standard IPC-4 benchmarks domain with axioms (both the version with axioms as well as the compiled version without axioms were provided in the IPC-4 benchmark set).
The task is to validate properties in systems of communicating processes (often communication protocols), encoded in the Promela language.
\citeauthor{edelkamp2003promela} (\citeyear{edelkamp2003promela}) developed an automatic translation from Promela into
PDDL, which was extended to generate the IPC domains.
We used the IPC-4 benchmarks for 
the Dining Philosophers problem (\pddl{philosophers}), and the  Optical Telegraph protocol (\pddl{optical-telegraph}).

\subsubsection{PSR}
The task of \pddl{PSR} (power supply restoration) is to reconfigure a faulty power distribution
network so as to resupply customers affected by the faults.
The network consists of electronic lines connected by switched and power sources with circuit breaker.
PSR was modeled as a planning problem in \citeauthor{bonet2003gpt} (\citeyear{bonet2003gpt}).
We used the version of PSR used in IPC-4, which is a simplified version of \citeauthor{bonet2003gpt} (\citeyear{bonet2003gpt}) with full observability of the world and no numeric optimization.

\begin{table}[htb]
	\centering
	\begin{tabular}{l|r|r|r|r}
		\hline
		2GB, 5min         
		& \#  
		& \rot{ASPlanS}
		& \rot{IPlanS}
		& \rot{Axioms} \\
		\hline
		\multicolumn{5}{c}{Domains from \citeauthor{ivankovic2015optimal} (\citeyear{ivankovic2015optimal})} \\
		\hline
		sokoban-axioms              & 30      & 5          & 4       & Y  \\
		sokoban-opt08-strips        & 30    & 4          & 0         & N\\
		trapping\_game              & 7       & 4          & N/A      & Y    \\
		\hline
		\multicolumn{5}{c}{Domains from \citeauthor{Ghosh2015} (\citeyear{Ghosh2015})} \\
		\hline
		acc-compiled                     & 8        & 0          & 0       &N  \\
		acc                         & 8        & 7          & 1       & Y  \\
		door-fixed                  & 2        & 1          & 1       & Y  \\
		door-broken                 & 2        & 0          & 0       & Y  \\
		grid-verification                  & 99       & 0          & 0       & Y  \\
		\hline
		\multicolumn{5}{c}{Domains from \citeauthor{kominis2015beliefs} (\citeyear{kominis2015beliefs})} \\
		\hline
		muddy-child-kg              & 7      & 1          &   N/A     & Y   \\
		muddy-children-kg           & 5      & 1          &   N/A     & Y   \\
		collab-and-comm-kg          & 3      & 1          &   N/A     & Y   \\
		wordrooms-kg                & 5      & 1          &   N/A     & Y   \\
		\hline
		\multicolumn{5}{c}{Domains from IPC} \\
		\hline
		psr-middle                  & 50     & 48         & N/A       & Y   \\
		psr-middle-noce             & 50     & 43         & 25       & Y \\
		psr-middle-compiled              & 50     & 1          & N/A       & N   \\
		psr-large                   & 50     & 21         & N/A       & Y   \\
		optical-telegraphs          & 48     & 0          & N/A        & Y  \\
		optical-telegraphs-compiled      & 48     & 0          & N/A        & N  \\
		philosophers                & 48     & 2          & N/A        & Y  \\
		philosophers-compiled            & 48     & 1          & N/A       & N  \\
		miconic & 150 & 29 & 25 & N \\
		miconic-axioms & 150 & 40 & 61 & Y \\
		grid & 5 & 2 & 0 & N \\
		grid-axioms & 5 & 3 & 2 & Y \\
		\hline
	\end{tabular}
	\caption{Results on domains with axioms. N/A indicates a lack of results since the current implementation of IPlan is not compatible with conditional effects. \# denotes the number of instances for each domain. The "Axioms" column indicates whether the corresponding domain has axioms}
	\label{tab:axioms}
\end{table}

\begin{figure}[htb]
\centering
	\includegraphics[width=\linewidth]{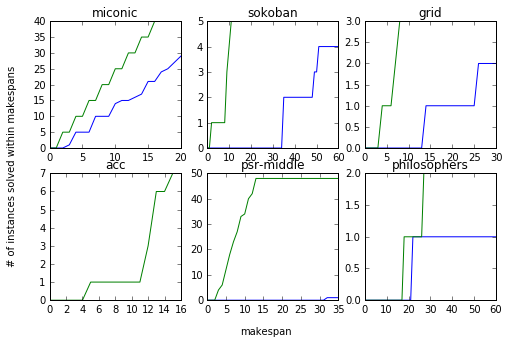}
	\caption{ASPlanS results showing the number of instances solved within makespan iterations (steps). The green lines represent domains with axioms, while the blue lines represent domains without axioms. Note that there can be instances which have solutions within a makespan but were not solved due to our time and memory limits.}
	\label{fig:asplan} 
\end{figure}

\begin{figure}[htb]
\centering
	\includegraphics[width=\linewidth]{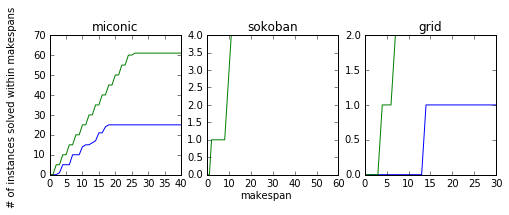}
	\caption{IPlanS results showing the number of instances solved within makespan iterations (steps). The green lines represent domains with axioms, while the blue lines represent domains without axioms. Note that there can be instances which have solutions  within a makespan but were  not solved due to our time and memory limits.}
	\label{fig:iplan} 
\end{figure}

\subsubsection{Grid and Miconic}
Like \pddl{Sokoban} some of the domains from existing IPC benchmarks without axioms can be reformulated to 
domains with axioms.
We chose \pddl{grid} and \pddl{miconic} from such domains for experimental evaluation.

In the \pddl{grid} domain,
the player walks around a maze to retrieve the key to
the goal. Just as in \pddl{Sokoban} the operators for the player's movement can be replaced with reachability axioms. Operators for retrieving keys or unlocking doors now have new reachability variables as their preconditions.

\pddl{miconic} is an elevator domain where we must transport a set of
passengers from their start floors to their destination floors. The up and down movements of an elevator can be expressed as axioms.

Note that in \pddl{Sokoban}, where \pddl{move} operators are zero cost, 
an optimal plan for an instance with axioms corresponds to an optimal plan for the original instance without axioms.
In \pddl{grid} and \pddl{miconic}, however, this no longer holds since the operators expressed as axioms have positive costs.

\subsubsection{Results}
The coverage results are shown in Table \ref{tab:axioms}.
In addition, Figures \ref{fig:asplan}-\ref{fig:iplan} show the number of instances solved within makespan iterations (steps).

In \pddl{Sokoban}, \pddl{acc}, \pddl{psr-middle}, \pddl{miconic}, \pddl{grid} and \pddl{philosophers} ASPlan (and sometimes IPlan) solved more instances
from the formulations with axioms than the compiled formulations without axioms.
This is because compiled instances tend to have longer makespans (as shown in Figure \ref{fig:asplan} and \ref{fig:iplan}), which tends to increase difficulty for model-based planners such as ASPlan and IPlan, because (1) the number of variables increases with makespan, and (2) in the iterative scheme used by model-basd planners (i.e., generating and attempting to solve $k$-step models with iteratively increasing $k$), problems with longer makespans increase the number of iterations which must be executed to find a plan.

\section{Conclusion}
\label{sec:conclusion}
We investigated the integration of PDDL derived predicates (axioms) into model-based planners.
We proposed axiom-aware model-based planners
ASPlan and IPlan.
ASPlan is an ASP-based planner, which is able to handle axioms and conditional effects.
IPlan is an IP-based planner based on Optiplan \cite{briel2005optiplan}.
We integrated axioms into IPlan using the ASP to IP translation method by \cite{liu2012answer}.
We evaluated ASPlan and IPlan on PDDL domains with axioms and showed that axiom-aware model-based planners can benefit from the fact that formulations with axioms tend to have shorter makespans than formulations of the same problem without axioms, resulting in higher coverage on the formulations with axioms.


\end{document}